\documentclass{vgtc}                          

\ifpdf
  \pdfoutput=1\relax                   
  \usepackage{graphicx}                
  \DeclareGraphicsExtensions{.pdf,.png,.jpg,.jpeg} 
\else
  \usepackage{graphicx}                
  \DeclareGraphicsExtensions{.eps}     
\fi%

\graphicspath{{figures/}{pictures/}{images/}{./}} 

\usepackage{microtype}                 
\PassOptionsToPackage{warn}{textcomp}  
\usepackage{textcomp}                  
\usepackage{mathptmx}                  
\usepackage{times}                     
\usepackage{cite}                      
\usepackage{tabu}                      
\usepackage{booktabs}                  

\onlineid{0}

\vgtccategory{Research}

\vgtcinsertpkg

\title{Immersive Rover Control and Obstacle Detection based on Extended Reality and Artificial Intelligence}

\author{Sofía Coloma*\\ %
        \parbox{1.4in}{\scriptsize SpaceR \centering \\ SnT, University of Luxembourg, Luxembourg.}
\and Alexandre Frantz*\\ %
        \parbox{1.4in}{\scriptsize SpaceR \centering \\ SnT, University of Luxembourg, Luxembourg.}
\and Dave van der Meer\\ %
        \parbox{1.4in}{\scriptsize SpaceR \centering \\ SnT, University of Luxembourg, Luxembourg.}
\and Ernest Skrzypczyk\\ %
        \parbox{1.4in}{\scriptsize SpaceR \centering \\ SnT, University of Luxembourg, Luxembourg.}
        \\
\and Andrej Orsula\\ %
        \parbox{1.4in}{\scriptsize SpaceR \centering \\ SnT, University of Luxembourg, Luxembourg.}
\and Miguel Olivares-Mendez\\ %
        \parbox{1.4in}{\scriptsize SpaceR \centering \\ SnT, University of Luxembourg, Luxembourg.}
}

\abstract{Lunar exploration has become a key focus, driving scientific and technological advances. Ongoing missions are deploying rovers to the Moon's surface, targeting the far side and south pole. However, these terrains pose challenges, emphasizing the need for precise obstacles and resource detection to avoid mission risks. This work proposes a novel system that integrates eXtended Reality (XR) and Artificial Intelligence (AI) to teleoperate lunar rovers. It is capable of autonomously detecting rocks and recreating an immersive 3D virtual environment of the robot's location. This system has been validated in a lunar laboratory to observe its advantages over traditional 2D-based teleoperation approaches.
} 

\CCScatlist{
  \CCScatTwelve{eXtended Reality}{Artificial Intelligence}{Space Robotics}{}
}

\begin{document}


\firstsection{Introduction}

\maketitle

In recent decades, space missions to the Moon have become increasingly relevant due to significant technological and scientific advancements, as well as humanity's goal of expanding to outer space. With growing interest from space agencies and private sectors, there is a demand to explore more hostile and unexplored environments with rovers, such as those located on the far side or south pole of the Moon. However, operating in such adverse terrain presents significant challenges, especially in identifying resources and obstacles, like rocks or formations, that may pose a risk to the mission. A small error, such as a collision with an undetected rock, can compromise not only the integrity of the rover but the entire mission. Traditionally, rover monitoring and teleoperation have been based on the interpretation of 2D images of the terrain along with the visualization of various rover parameters and environmental data\cite{schreckenghost2009measuring}. However, depending on the scenario, this system may not provide enough detail or intuition to prevent accidents or identify objects of interest accurately. In this context, it is advisable to equip rovers with advanced technologies to ensure safety and success in future missions, aiming to monitor and control the rovers from closer locations, e.g., in the Lunar Gateway or Lunar Base \cite{edwards2021moon,burns2019science}, where the latency will be lower than from Earth.

In particular, the integration of more immersive and autonomous systems significantly enhances the operator's ability to effortlessly identify objects, particularly in challenging low-light conditions that often make objects in 2D images difficult to distinguish. Some advanced technologies not only increase obstacle detection and management capabilities but can also enable a 3D representation of the lunar environment, improving depth perception, spatial relationships, and terrain understanding compared to a 2D representation. With these advances, the way is paved for more efficient and safer lunar exploration, opening new possibilities for scientific research and technological developments in space. Thereby, to contribute to the ever-advancing space sector, the presented work proposes a novel system to teleoperate rovers in unknown and hostile environments, able to detect relevant obstacles or resources on the lunar surface. The application combines Extended Reality (XR) with Artificial Intelligence (AI) models with the aim of effectively detecting and visualizing rocks in 3D virtual environments. This integration not only represents a breakthrough in autonomous rock detection and identification in unknown environments but also immerses operators in a detailed and life-like experience. Thus, it increases the ease of operator decision-making and improves the effectiveness of exploration operations.

\section{Autonomous object detection and localization for immersive XR rover teleoperation}
\subsection{Architecture}
\begin{figure}[ht]
    \centering
    \includegraphics[width=0.99\linewidth, height=5cm]{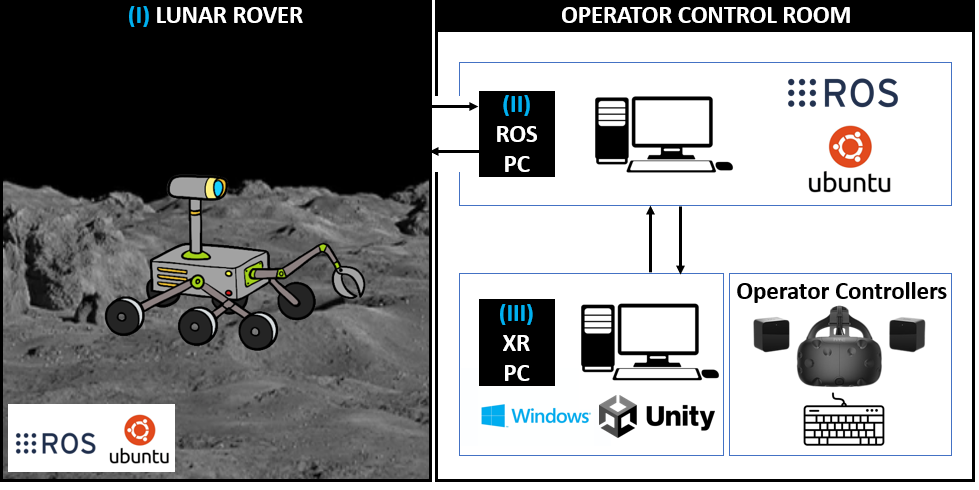}
    \caption{Architecture of the XR system connected to the lunar rover.}
    \label{fig:architecture}
\end{figure}

As Figure \ref{fig:architecture} illustrates, the system consists of three main subsystems: (I) the lunar rover, (II) the ROS PC, and (III) the XR PC. The architecture facilitates the transmission of data from the rover's internal camera and sensors to the ROS PC for processing, including camera feed and point cloud information. The processed data is used in the PC XR to detect and label rocks in a 3D environment. The XR PC is connected to enable data transfer and can send teleoperation commands to the lunar rover, allowing the operator to use a Human Interface Device (HID) in the control room, such as a keyboard or gamepad, and more importantly, the HTC VIVE Cosmos Elite Virtual Reality (VR) system to interact with the rover and the environment.

Regarding the network, we use a bridge connection, i.e. a dedicated ROS bridge server, that is employed between the ROS and XR PCs. In terms of Operating Systems (OS), both GNU/Linux and Windows are used. The lunar rover and the ROS PC run Ubuntu with ROS, and the XR PC runs the Unity3D game engine on Windows. For this work, Unity3D provides the necessary processing power, and through bridge connection and dedicated Unity3D packages, we achieve interoperability between ROS and Unity3D processes. This system architecture allows the XR system to run on Windows while processing ROS data directly in Unity3D.

\subsection{Phases}
The system detailed in the presented work is structured into three main phases, each of which plays a critical role in ensuring a logical and smooth progression.
\begin{itemize}
\item In the first phase, sensory data from the remote rover's sensors, including RGBD information, colour images and depth data from the Intel RealSense D-455 camera, are collected during the mission. 

\item The second phase involves processing this data, utilizing the YOLOv5 CNN algorithm \cite{jocher2022ultralytics} to detect rocks in 2D images, and generating a 3D mesh of the rover's surroundings based on the RTAB-MAP\cite{labbe2019rtab} and 3D point cloud environment generation.

\item The third phase visualizes the processed data in a 3D reconstructed XR environment, providing the operator with a dynamic view of the rover's surroundings. It includes a 3D model of the rover and marked 3D visual indicators for identified rocks and their position. 

\end{itemize}
This comprehensive approach enhances the operator's ability to discern the positions of rocks proximate to the rover in a lunar environment. 

\section{Experiments and results}
\begin{figure}[ht]
    \centering
    \includegraphics[width=0.99\linewidth, height=6cm]
    {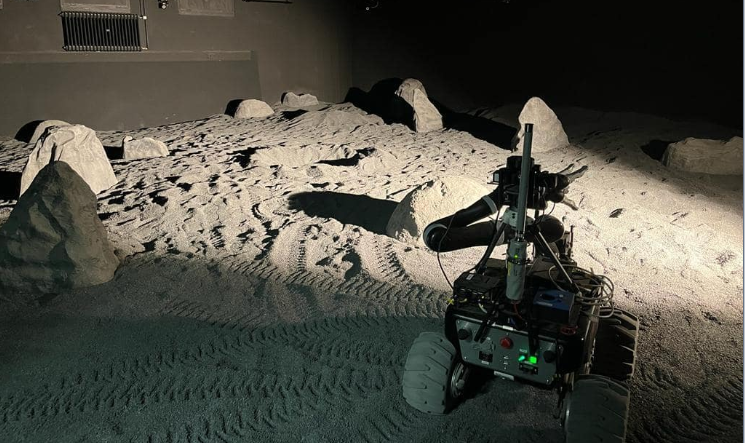}
    \caption{Lunar rover teleoperated at the LunaLab to avoid the rock obstacles during the test.}
    \label{fig:architecture1}
\end{figure}

To check the robustness and applicability of the work developed, a series of tests were carried out in an analogue lunar laboratory, the LunaLab, at the University of Luxembourg \cite{coloma2022enhancing}. In particular, the system was validated with two experiments and five participants using the presented architecture and configuration of Figure \ref{fig:architecture}. Both experiments consisted of teleoperating a rover at the LunaLab, where the participants from a remote control room had to detect and avoid colliding with the rocks located in the terrain, see Figure \ref{fig:architecture1}. In one of the tests, the user-controlled the rover's movements based solely on visual information received from the rover's integrated camera. This configuration involved relying on RGB images displayed on a computer screen. Instead, to contrast teleoperation methods involving 2D versus 3D, the operator controlled the rover using XR in the other test. In this case, the operator visualized the environment in the form of a 3D mesh using the Head Mounted Display (HMD).

Through a study with different users, the performance and advantages of our XR system were contrasted with traditional teleoperation approaches that are based on 2D image interpretation. The findings derived from these experiments support the relevance and significant contributions that this work brings to the field of rover teleoperation and lunar exploration using XR. The implementation of the system in XR has not only shown a significant impact in minimizing the cognitive load of the operators in complex areas with obstacles, but also, participants reported a greater perception of the environment while using the XR system. However, the importance of this system and the validation carried out extends beyond lunar exploration. It can be successfully applied to various contexts on Earth, providing innovative and advanced solutions. Examples include scenarios such as search and rescue operations or natural disaster exploration.


\section{Conclusion and future work}
This work has presented and validated a novel rover teleoperation system using XR in a lunar environment, highlighting the importance of AI in the identification and management of obstacles. The implementation of the XR system has proven to reduce the cognitive load of operators significantly. Considering the future, it is essential to expand the range and functionality of the XR system to include enhanced ranging and reporting capabilities in the virtual lunar environment. Accurate integration of distance metrics and real-time alerts within the XR system will not only increase the rover's safety when navigating challenging terrain. It will also provide operators with critical data to make informed operational adjustments and optimize exploration routes. Therefore, future work should focus on developing and validating algorithms that allow distances and obstacles to be accurately measured in the virtual environment, as well as how that information is intuitively transmitted and displayed to operators in the XR interface.

\bibliographystyle{abbrv-doi}

\bibliography{template}
\end{document}